\documentclass[journal]{IEEEtran}

\usepackage{algorithm}
\usepackage{algorithmic}
\usepackage{cite}
\usepackage{amsmath,amssymb,amsfonts}
\usepackage{algorithmic}
\usepackage{graphicx}
\usepackage{textcomp}
\usepackage{xcolor}
\usepackage{subfigure}
\usepackage{bm}
\allowdisplaybreaks[4]
\def\BibTeX{{\rm B\kern-.05em{\sc i\kern-.025em b}\kern-.08em
    T\kern-.1667em\lower.7ex\hbox{E}\kern-.125emX}}

\begin{document}

\title{Equitable Time-Varying Pricing Tariff Design: \\ A Joint Learning and Optimization Approach}
%
%
%

\author{Liudong~Chen
        and~Bolun~Xu
}


\maketitle

\begin{abstract}
Time-varying pricing tariffs incentivize consumers to shift their electricity demand and reduce costs, but may increase the energy burden for consumers with limited response capability. The utility must thus balance affordability and response incentives when designing these tariffs by considering consumers' response expectations. This paper proposes a joint learning-based identification and optimization method to design equitable time-varying tariffs. Our proposed method encodes historical prices and demand response data into a recurrent neural network (RNN) to capture high-dimensional and non-linear consumer price response behaviors. We then embed the RNN into the tariff design optimization, formulating a non-linear optimization problem with a quadratic objective. We propose a gradient-based solution method that achieves fast and scalable computation. Simulation using real-world consumer data shows that our equitable tariffs protect low-income consumers from price surges while effectively motivating consumers to reduce peak demand. The method also ensures revenue recovery for the utility company and achieves robust performance against demand response uncertainties and prediction errors.



\end{abstract}


%
\IEEEpeerreviewmaketitle

\section{Introduction}
Electricity demand will continue to grow due to electrification and increasing occurrences of weather extremes~\cite{energy}. Concurrently, consumers are  becoming more responsive to electricity prices due to the increasing adoption of smart and flexible resources such as home energy storage, smart heat pumps, and home energy management systems~\cite{qdr2006benefits, el2020contracted}. Increasing peak demand significantly stresses the distribution grid security and generation capacity. To this end, utilities are adopting time-varying pricing tariffs, including time of use (ToU) tariffs, and hourly different pricing scheme~\cite{review}, to incentivize consumers to shift their electricity demand and reduce costs. 

ToU tariffs allow operators to take predefined non-flat prices over a long time horizon, which provides consumers the possibility to arrange demand schedules. Utilities worldwide have employed ToU tariffs to facilitate demand shifting, such as incentivizing electric vehicles (EV) to charge during off-peak periods~\cite{tou3, tou9, tou4, tou5}. However, as more consumers become price-responsive, ToU may lead to new demand peaks and fail its design objective to flatten demand, as consumers will simultaneously increase their electricity consumption at the low price periods~\cite{toupeak}.
Besides, while ToU issues different prices over the day, these prices are fixed throughout the seasons. ToU hence is ostensibly a blunt pricing tool to regulate consumers' demand~\cite{tou8} in response to changing grid conditions, especially as the renewable penetration and the share of distributed energy resources (DER) increase.

Utilities are now experimenting with dynamic time-varying tariffs beyond ToU to more frequently update price schedules and to better align consumers' responses to changing grid conditions~\cite{rtp5, rtp6}. These prices should jointly reflect wholesale market prices,  distribution grid security constraints, and consumers' willingness to response. On the other hand, utilities must also address energy equity as the volatility in dynamic price tariffs may significantly increase the cost of electricity for low-income consumers who lack the means to reduce demand during peak price periods~\cite{review, nyt}. 
To address these challenges, we provide a novel equitable solution to design day-ahead pricing tariffs using a joint learning and optimization approach. This method accurately capture the price response behavior of individual consumers and design tiered prices consider their social demographics. The detailed contributions are as follows:
\begin{enumerate}
    \item 
    We use energy burden to measure energy equity and formulate the equitable tariff design problem as a non-linear optimization problem, including an embedded recurrent neural network (RNN) architecture for modeling consumers' time-coupling and nonlinear responses to time-varying prices. 
    
    \item 
    We propose a joint learning-based identification and tariff optimization approach by training the RNN using historical price and demand measurements to develop up-to-date consumer response models, then directly incorporating the trained RNN into tariff optimization.  
    
    \item We develop a gradient-based algorithm to solve the optimization problem with proven convergence performance and computation scalability. 
    
    \item 
    We design a case study based on real-world data using the proposed approach to design income-tier-based tariffs by modulating wholesale day-ahead market prices. Our approach can effectively achieve peak shaving functions without adding energy burdens to low-income consumers. 
\end{enumerate}

The remaining of the paper is organized as follows: Section~II reviews related literature, Section~III and~IV introduces the proposed equitable tariffs design model and solution algorithm, Section~V describe simulation results under demand response (DR) and price surge settings, and Section~VI concludes the paper.




\section{Backgrounds and Literature Review}
\subsection{Time-varying tariffs}

Issuing consumers with different prices throughout the day is a convenient way to incentivize demand responses for aligning utilities' needs with consumers' cost-saving motivations~\cite{review}. 
ToU tariffs provide fixed daily price schedules monthly and have been widely adopted by utilities to incentivize consumer responses. For example, Con Edison has a ToU plan with peak and off-peak rates for residential, business, and EVs~\cite{tou4}. The Salt River Project also uses a ToU pricing plan to provide consumers with the greatest opportunity for savings by directing their energy use towards off-peak hours~\cite{tou5}. With ToU tariffs, consumers can arrange their schedules to shift or reduce demand, and their demand adjustment behaviors are predictable, helping operators regulate demand distributions effectively~\cite{tou1}. Many methods are designed to calculate ToU tariffs, including the Gaussian mixture model clustering technique~\cite{tou6} and combinations of statistical analysis and optimization algorithms~\cite{tou7}. 

Utilities are now experimenting with more advanced time-varying tariffs beyond the ToU, such as day-ahead tariffs, which provide consumers with a different price schedule every day~\cite{rtp5, rtp6}. This effort aims to update tariffs to align with the changing grid conditions, such as distribution grids~\cite{tou3} or wholesale markets~\cite{tou8, rtp2, rtp3, rtp4}. 
Many studies also focus on the design and application of time-varying pricing tariffs. These design methods include a game-theoretic approach with consumers' occupancy status and consumption patterns\cite{rtp7}, a dynamical pricing model based on the supply and demand ratio~\cite{rtp8}, and an online learning method~\cite{rtp9}. Time-varying tariffs are usually used in designing the billing strategy of EV charging processes~\cite{rtp10}, helping consumers control their loads while reducing electricity expenditures~\cite{rtp11}.

\subsection{Equity and behavior models in utility tariff designs}
Consumers' demand flexibility is highly associated with their socio-demographic information, such as job type, income level, and household size~\cite{ATUS}. In general, low-income consumers are less flexible in responding due to more restricted working schedules and less access to flexible demand technologies such as smart home controllers or home energy storage. Meaning time-varying tariffs may increase their energy burden as low-income consumers are less likely to reduce their demand during high-price periods~\cite{rtp4}. Additionally, the adoption of solar photovoltaic (PV) by high-income households, which has increased with the development of DERs, will also increase costs for low-income consumers without PV~\cite{add1}. Thus, power systems must consider energy equity and design new tariffs in a fair and effective manner, which requires accurate consideration of consumers' response behaviors~\cite{add4}..


Modeling consumers' price response is challenging as their behavior is subjective and often nonlinear features~\cite{Nat1}. Researchers use social science methods to capture user behavior, such as multi-criterion decision-making based on a brief survey to determine the weight of different objectives in DR optimization~\cite{add4}, and prospect theory to mimic consumers' decision-making preferences when facing risks~\cite{Chen}. However, these methods' computational complexity and non-differentiability features make it difficult to calculate behaviors and unsuitable for efficient gradient-based optimization required for various applications. Additionally, consumers' price response is time-coupled, where load consumption shifts to different time slots in response to price signals, presenting another challenge in identifying behaviors.

\subsection{Machine learning for DER management}
Machine learning techniques emerged as an effective way to model diverse and nonlinear participating and response behaviors in electricity markets with partial knowledge. For instance, neural networks can represent the lost load market clearing problem in the bi-level optimal bidding model~\cite{ml1} and capture the power system operation characteristics to solve the optimal power flow problem~\cite{ml4}. Additionally, in building energy systems with complex electrical and thermal dynamics, neural networks are also an emerging useful tool to capture system dynamics and devise optimal control schemes for energy consumption, which opens the door to incorporate system dynamics with decision-making~\cite{ml2, ml3}. In contrast to previous work, we propose a data-driven approach to design income-tiered day-ahead time-varying pricing tariffs. Using an RNN learning model, we can accurately capture time-coupled consumers' price response behavior by observing their historical response to price signals, providing a practical framework for developing equitable time-varying tariffs to facilitate demand flexibility.

\section{Model and Formulation}

\begin{figure}[htbp]
    \centering{
    \includegraphics[width=0.48\textwidth]{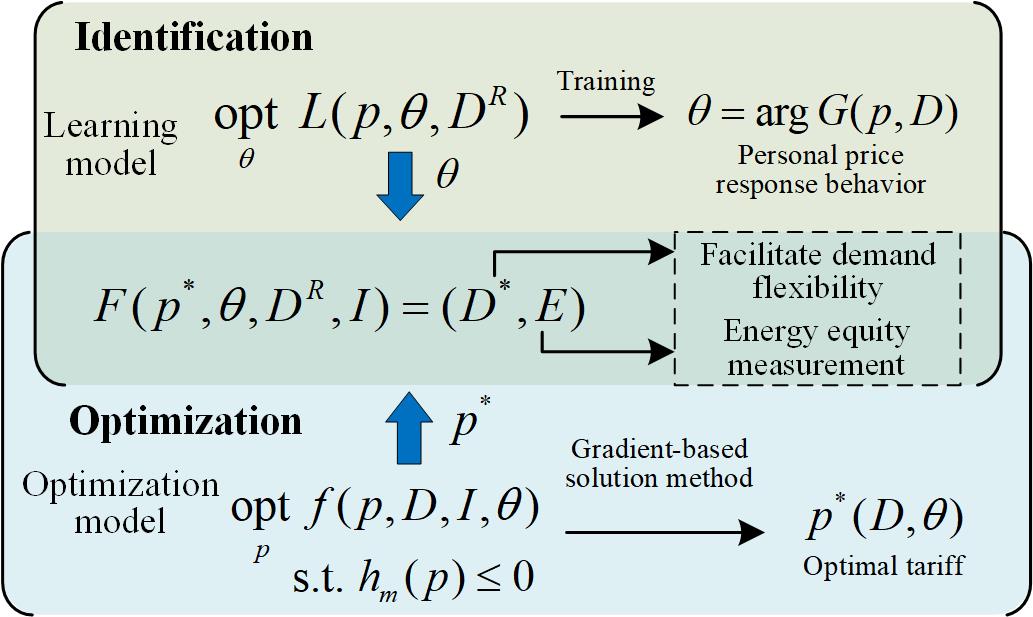}}
    \caption{Joint identification and optimization method}
    \label{fig1}
    \vspace{-.2cm}
\end{figure}

We consider a scenario in which a utility supplying consumers with income-based time-varying price tariffs to incentivize response from demands while also controlling the induced energy burden over low-income consumers. The utility designs and issues a new price schedule each day ahead based on wholesale market clearing prices through a joint optimization and learning framework as shown in Fig. \ref{fig1}. Through the framework, we can obtain the price response behavior, i.e., relationships between prices $p$ and demands $D$ with their price response parameters $\theta$ and social demographics $I$, which then generates affordable and effective time-varying electricity tariffs for each consumer, i.e., suitable for their income tier and price response behaviors, despite the tariff volatility, and enable utilities to facilitate demand-side flexibility and realize energy equity.

We formulate a nonlinear optimization problem including a quadratic objective, utilities' revenue recovery constraint, and consumers' response modeled using RNN. The overall goal of the tariff is to redistribute consumers' electricity bills and protect the interests of low-income consumers when participating in price response. We optimize $t \in T$ periods (24 hours in our case studies) over $\mathcal{I}$ consumer groups following $m\in M$ constraints. $\mathcal{I}_n$ represents energy burden-based consumer sub-groups, i.e., consumers within the same sub-group have similar energy burden.

The \textbf{optimization} model designs the price tariff schedule $\bm{p}_i$ for each consumer based on the updated RNN model $G_i(\bm{p}_i|\bm{\theta}_i)$ which predicts consumers' DR based on the price tariff
\begin{subequations}
\begin{align}
    \min_{\bm{p}_i} f=& \sum_{i\in\mathcal{I}} \Big(\Big[\frac{\bm{D}_i^T \bm{p}_i}{I_i} - \overline{E}\Big]^+\Big)^2 + \alpha \lVert \bm{p}_{i} - \bm{\lambda} \rVert_{2}^{2} \label{obj}\\
    &\text{s.t.}  \nonumber\\
    &\bm{D}_i = G_i(\bm{p}_i|\bm{\theta}_i) \label{dr} \\
    &\textstyle \sum_{i\in\mathcal{I}} \bm{D}_i^T \bm{p}_i \geq C + \bm{D}_{0,i}^T \bm{\lambda} \label{rev}\\
    &\bm{p}_i  = \bm{p}_j, \; \forall i,j\in \mathcal{I}_n, \forall n\in \mathcal{N} \label{group}
\end{align}
\end{subequations}
where $[ x]^+ = \max\{0,x\}$ is the positive value function. The objective function reduces the energy burden of low-income consumers by modulating the tariff price from the reference wholesale market price $\bm{\lambda}$.
$\frac{\bm{D}_i^T \bm{p}_i}{I_i}=E_i$ is the energy burden of consumer $i$~\cite{report_energyburden}, in which $\bm{p}_{i}$ and $ \bm{D}_{i}$ are the pricing and demand profile for each consumer $i$, and $I_i$ is the income data. $\bm{D}_{0,i}$ is the baseline demand profile when receiving the wholesale market prices. $\overline{E}$ is the desired energy burden threshold. $\alpha$ is the weighting factor for the wholesale price modulation.

\eqref{dr} uses a RNN $G_i$, parameterized with $\bm{\theta}_i$, to represent each consumer's price response behavior. \eqref{rev} is the utility company's revenue adequacy constraint ensuring the payment collected from the consumer affords the wholesale market purchasing cost and a constant O\&M cost $C$. \eqref{group} ensures all consumers from the same group receive the same price. 
The energy burden includes consumers' income information and energy consumption. Grouping consumers by energy burden promotes fairness by assigning similar energy burden groups the same electricity price. Besides, in practice, due to the large number of consumers and similarities in their performance, it is unnecessary to create personalized tariffs for each individual. While consumers in each group receive the same tariff, they still respond to the tariffs based on their individual behavior and income level, allowing them to be considered equitable entities.

Our model also includes an optional demand reduction constraint if the utility wishes to incorporate a demand reduction event in the tariff:
\begin{align}
    \sum_{i \in \mathcal{I}} D_{i,t} \leq (1-\beta) \sum_{i \in \mathcal{I}} D_{\mathrm{0},i,t}, \ \forall t \in \{ \mathrm{PeakHours} \} \label{demand2}
\end{align}
where $\beta$ is the required demand reduction ratio in peak hours, and $\{\mathrm{PeakHours}\}$ is the set of hours determined by the utility to conduct demand response.

The optimization model is associated with the following \textbf{identification} model, which trains and updates the RNN model parameters using past price data and observed consumer demand $\bm{D}_i^{R}$ in response to the instructed prices:
\begin{align}
&\min \limits_{\bm{\theta}} L=\lVert G_i(\bm{p}_i|\bm{\theta})- \bm{D}_i^{R} \rVert_{2}^{2}
\end{align}

Hence, the utility executes the identification and the optimization model alternatively to update the user behavior RNN model based on past prices and responses, and use the updated model to design new tariffs.

\section{Solution Methods}

\subsection{Price responses model structure design and training}

We initially design a network structure for the identification process, as illustrated in Fig. \ref{fig2}. The $G$ block consists of several layers: an input layer with $T$ neurons representing prices in each time slot, where the number of neurons is determined by the length of the price time series; multiple hidden layers with varying numbers of neurons, which are determined based on the data characteristics; and an output layer with $T$ neurons corresponding to the demand change. The time series of demand determines the number of output neurons. Notably, we focus on learning the demand change during the DR period, denoted as $\Delta D$, because the demand change constitutes only a small portion of the baseline demand. Learning the baseline demand directly proves challenging in adequately capturing the demand change. By utilizing this network structure formulation, we can effectively model the consumer's price response behavior over $T$ hours.
Besides, more blocks can be constructed in the identification process, e.g., noise blocks formulate the agent prediction error (difference between real price response results and agent model results). Here we only consider the price response behavior block. Then the network and neurons model for one consumer can be defined as:

\begin{figure}[htbp]
    \centering{
    \includegraphics[width=0.48\textwidth]{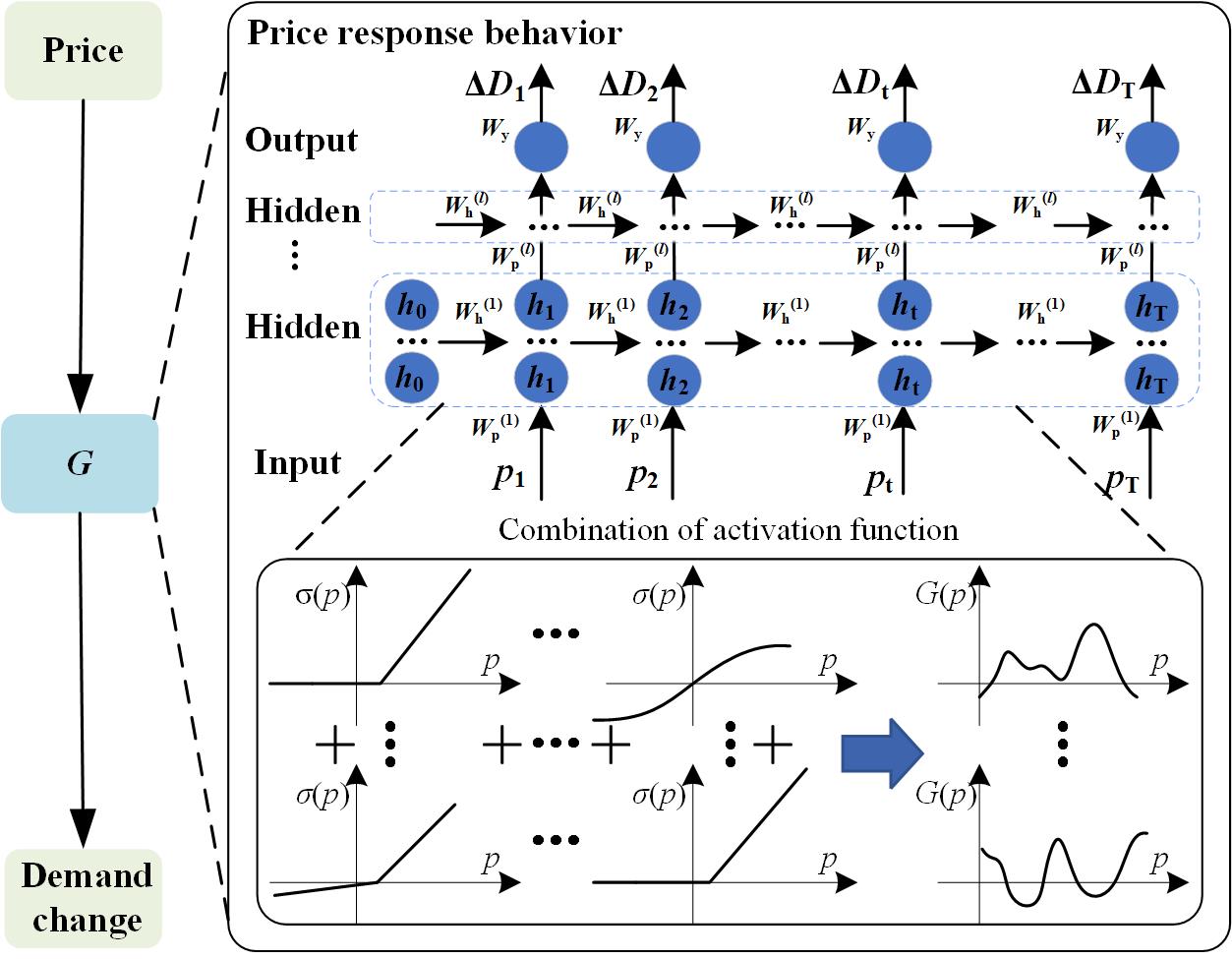}}
    \caption{Consumers' price response behavior identification}
    \label{fig2}
    \vspace{-.2cm}
\end{figure}
\begin{align}
\Delta D_{t} &= G \left( p_{t}, \bm{\theta} \right) \label{rnn1}\\
\bm{\theta}&=\{ \bm{W}_{\mathrm{p}}, \bm{W}_{\mathrm{h}}, \bm{W}_{\mathrm{y}}, \bm{b}_{\mathrm{h}}, b_{\mathrm{y}} \}\\
\bm{h}_{t}^{(1)}&=\sigma(p_{t} \bm{W}_{\mathrm{p}}^{(1)} 
+ \bm{h}_{t-1}^{(1)} \bm{W}_{\mathrm{h}}^{(1)} 
+ \bm{b}_{\mathrm{h}}^{(1)}) \label{rnn2}\\
\bm{h}_{t}^{(l)}&=\sigma(\bm{h}_{t}^{(l-1)} \bm{W}_{\mathrm{p}}^{(l)} 
+ \bm{h}_{t-1}^{(l)} \bm{W}_{\mathrm{h}}^{(l)}  
+ \bm{b}_{\mathrm{h}}^{(l)} ) \label{rnn3}\\
\Delta D_{t}&= \sigma(\bm{h}_{t}^{(L)} \bm{W}_{\mathrm{y}}+b_{\mathrm{y}}) \label{rnn4}\\
\bm{h}_{0}^{(l)}&=\bm{0}, \   \forall l \in L \label{rnn5}
\end{align}
where $l$ is the hidden layer index of the network (except input and output layer), and $l \in L$. $\bm{h}_{t}$ is the vector of the hidden variable in the hidden layers, whose dimension is determined by the neurons of the corresponding hidden layer. $\bm{W}_{\mathrm{p}}, \bm{W}_{\mathrm{h}}, \bm{W}_{\mathrm{y}}$ are network weighting parameters of input feature $p$, hidden variable $\bm{h}_{t-1}$, and hidden variable in output layer $\bm{h}_{t}$, the dimension of the matrices corresponding to the number of neurons; $\bm{b}_{\mathrm{h}}, b_{\mathrm{y}}$ are the network bias parameters in the hidden layers and output layers, $\bm{b}_{\mathrm{h}}$ is a vector with the same dimension of the neurons number in the hidden layers, and $b_{\mathrm{y}}$ is a number. $\sigma(\cdot)$ is the activation function used to formulate non-linear features of the network, which may differ in layers.

The input to this network is the time series of prices, and the learning process takes place per time period. After processing the first time slot, the learning results are recorded through hidden variables and subsequently updated at each new time slot based on the new time slot's features. This process enable the time series features of prices transfer in the network training process, and realizes the time coupled relationship learning with the demand change. Besides, the network share the same parameters $\bm{W,b}$ in all time slots, which reduces the computation burden of the network.

To train the network, we use an agent model to simulate consumers' price response process, which explicitly describe in Appendix. Consumer will change their reduced demand and shiftable demand to minimize their electricity bill response to specific tariff patterns. Then we can obtain the demand change data and corresponding price data as the output label and input feature of the RNN, respectively, which can be expressed as follows:
\begin{align}
&\{[\bm{p}_{i}],[\bm{\Delta D}_i^{R}]\}
\end{align}

With the datast, we are able to train the RNN and get the parameters $\bm{\theta}$. Here the RNN training is followed by the forward calculation and backward propagation. The gradient of loss function $L$ with regards to $\bm{\theta}$ is calculated by derivative chain rule and computational graphs layer by layer, and the learning rate is computed by adaptive moment estimation (Adam)~\cite{mitbook}.

\subsection{Gradient-based optimization solution algorithm}
The tariff optimization is a non-convex problem due to the inclusion of RNN for modeling consumer responses. 
We introduce a gradient-based solution algorithm including an identification stage and an optimization stage. 
To incorporate constraints into the gradient-based algorithm, we move the constraints to the objective function using log-barrier functions to keep the decision variables within bounds during iterations. The augmented optimization problem can be expressed as:
\begin{align}
\begin{split}
        \min \limits_{\bm{p}_i(\mu)}\ &F_{0}=\mu f+\sum_{m \in M} \varphi_{m}(C) \\
        &= \mu  \Big( \sum_{i\in\mathcal{I}} \Big(\Big[E_{i} - \overline{E}\Big]^+\Big)^2 + \alpha \lVert \bm{p}_{i} - \bm{\lambda} \rVert_{2}^{2}   \Big)\\
    &+ \sum_{t\in \{\mathrm{PK}\}} \ln(C_{t}) + \ln(C_{1}) \label{newobj}
\end{split}   
\end{align}
 \vspace{-0.2cm}
\begin{align}
        &C_{t}=-\sum_{i\in\mathcal{I}} D_{i,t}+(1-\beta)* \sum_{i\in\mathcal{I}} D_{\mathrm{0},i,t}, \ \forall t\in \{\mathrm{PeakHours}\} \\
        &C_{1}=-\sum_{i\in\mathcal{I}} \bm{D}_i^T \bm{p}_i + \bm{D}_{0,i}^T \bm{\lambda}
\end{align}
where $\varphi$ is the penalty function, and $\varphi(C)=\ln(C)$; $\mu$ is the parameter of the algorithm, which increase by iterations. We consider constraints in the gradient of the objective function, and the gradient of $F_{0}$ with regard to $\bm{p}_i(\mu)$ will be expressed as a function of parameter $\mu$. 

With the parameters $\bm{W,b}$ from the identification process, we are able to calculate the gradient of $F_{0}$ with regard to $p_{i,t}$, which is also obtained through the derivative chain rule: 
\begin{align}
\begin{split}
      \frac{\partial	F_{0}}{\partial p_{i,t}}&=\frac{\partial	F_{0}}{\partial E_{i}}  \left( \frac{\partial E_{i}}{\partial D_{i,t}}  \frac{\partial	D_{i,t}}{\partial p_{i,t}} 
      + \frac{\partial E_{i}}{\partial p_{i,t}}  \right)  + \sum_{m} \frac{\partial \varphi_{m}}{\partial p_{i,t}}\\
      &= \frac{2\mu (E_{i}-E_{\mathrm{bond}})}{I_{i}} 
       \left( D_{i,t} + \frac{\partial D_{i,t}  p_{i,t}}{\partial p_{i,t}} \right) \\
      &-\delta(t)   \frac{\partial D_{i,t}}{\partial p_{i,t}} \frac{1}{{C}_{t}} 
      -\left(D_{i,t} + \frac{\partial D_{i,t}p_{i,t}}{\partial p_{i,t}} \right)\frac{1}{{C}_{1}}  \label{der}  
\end{split}
\end{align}
\begin{align}
    \delta(t)=\left \{
        \begin{aligned}
         &1, \ \ \mathrm{if} \  {t\in \{\mathrm{PeakHours}\}} \\
         &0, \ \ \mathrm{OTW}
        \end{aligned}
    \right.
\end{align}

The most critical part is to calculate the partial derivative $\frac{\partial	D_{i,t}}{\partial p_{i,t}}$, which follows the RNN architecture and should be obtained through derivative chain rule and computational graphs. It is noted that the demand change of consumer $i$ in time slot $t$ is the output of neuron $t$ in the output layer. Here we only analyze one neuron for example.
\begin{equation}
\begin{split}
    \frac{\partial D_{t}}{\partial p_{i,t}} &= \frac{\partial D_{t}}{\partial \bm{h}_{t}^{(L)}}  
     \prod_{l=2}^{L} \left( \frac{\partial \bm{h}_{t}^{(l)}}{\partial \bm{h}_{t}^{(l-1)}} 
     + \frac{\partial \bm{h}_{t}^{(l)}}{\partial \bm{h}_{t-1}^{(l)}} 
    \right)
    \left( \frac{\partial \bm{h}_{t}^{(1)}}{\partial p_{i,t}} + \frac{\partial \bm{h}_{t}^{(1)}}{\partial \bm{h}_{t-1}^{(1)}} \right) \\
    &=\bm{W}_{y} \frac{\partial \sigma}{\partial \bm{h}_{t}^{(L)}}
     \prod_{l=2}^{L} \left( \frac{\partial \sigma}{\partial \bm{h}_{t}^{(l-1)}}  \bm{W}_{p}^{(l)} 
     + \frac{\partial \sigma}{\partial \bm{h}_{t-1}^{(l)}}  \bm{W}_{h}^{(l)}
    \right) \\
    &*\left( \frac{\partial \sigma}{\partial {p}_{i,t}} \bm{W}_{p}^{(1)} + \frac{\partial \sigma}{\partial \bm{h}_{t-1}^{(1)}} \bm{W}_{h}^{(1)} \right) \label{derend}
\end{split}
\end{equation}

Since each consumer corresponds to a different structured network, this step should be repeated in each network to obtain all partial derivatives in (\ref{der}). After getting the gradient of the objective function, we use Hessian matrices to calculate the Newton step and update the decision variables during the iteration process. 

Previous work and practices have demonstrated that gradient-based algorithms can achieve good convergence performance in RNN~\cite{globalmini, proof2}. Using second-order gradient-based optimization methods can capture curvature information, enabling them to seamlessly escape flat regions and saddle points and contribute to fast convergence and high robustness~\cite{proof4}.
These demonstrate that the identification (network training) and optimization process related to RNN architecture can converge effectively. In the optimization process, the variables are tariff $p$, whose number is significantly less than the weighting variables $\bm{W}$ in the training process. Besides, the gradient of these two variables has the same structure as expressed in the derivative chain and computation graph (Fig. \ref{fig2}). Specifically, according to (\ref{rnn2})-(\ref{rnn3}), taking derivative with regards to $p$ or $\bm{W}$ will get similar results with the same structure. 
Thus, using gradient-based solution algorithms can at least converge to a local minimum that satisfies the error requirements (Please see detailed proof in Appendix). In the case study, we also simulate many times to show the robustness of the solution results. 


\begin{algorithm}
	\renewcommand{\algorithmicrequire}{\textbf{Input:}}
	\renewcommand{\algorithmicensure}{\textbf{Output:}}
        \caption{Gradient-based solution algorithm} 
	\label{alg3} 
	\begin{algorithmic}[1]

  \STATE  Initialization: Dataset $\{\bm{p}_i,\bm{\Delta D}_i^{R}\}$, parameters $\mu$, $\bm{\lambda_{i}}$, $\bm{D}_{0,i}$, $\alpha$, $\beta$, $I_{i}$, $\overline{E}$, tolerance $\epsilon=1e^{-6}$, number of constraints $M$.
	\STATE Train RNN using dataset, obtaining price response function $\bm{D}_{n}=G_{n}(\bm{p}_{n}|\bm{\theta}_{n})$ for each group. 

		\WHILE{$M / \mu  \geq \epsilon$} 
                \WHILE{$ \Vert {\nabla \bm{p}_n} \Vert \geq \epsilon$}
		      \STATE Calculate derivative $\nabla \bm{p}_{n}$ of $F_{0}(\bm{p}_{n})$ based on (\ref{der})-(\ref{derend})
                \STATE Choose step size $v$ via backtracking line search
		      \STATE Update $\bm{p}_{n}=\bm{p}_{n}+v \Delta \bm{p}_{n}$ 
		\ENDWHILE
		\STATE Update $\bm{p}_n=\mu \bm{p}_{n}$
            \STATE Increase $\mu=10\mu$
		\ENDWHILE 
         \STATE Broadcast modified tariffs $\bm{p}_{n}^*$ to consumers, obtain consumers modified demand $\bm{D}_{i}^*$ by agent model.
        \ENSURE Day-ahead time varying electricity tariff $\bm{p}_{n}^*$ and modified consumers demand $\bm{D}_{i}^*$
	\end{algorithmic} 
 \end{algorithm}

\section{Simulation}
 In this section, we first provide consumers' price response identification results. Then we design three case studies to analyze the equity time-varying tariffs design results: (1) tariff design; (2) DR events with demand reduction target; (3) price surge with extremely high wholesale market prices.

\subsection{Data}
In this study, we use the raw data of Austin, Texas, from ERCOT, which includes day-ahead 15-minute granularity price and corresponding load consumption data, including 25 consumers in 2018. Here we use one-day load consumption to represent one consumer and transfer 15-minute granularity price and load consumption data to hourly data. Then we choose 40 days of hourly load consumption data of 25 consumers, and calculate their demand change during price response processes using the agent model. We build the dataset with 1000 consumers and 1460 historical day-ahead electricity prices and corresponding demand change information through the overall formulation. In the optimization process, 1000 consumers' incomes are randomly generated between 800\$ to 60000\$ and divided over 365 days to get their daily income. Then we rank consumers according to their energy burden and separate consumers into 10 energy burden groups with 100 consumers in each group and use consumers' average value to express the performance of groups. The parameters $\overline{E}$ is set as 0.06, $\alpha$ is set as 1.

\subsection{Consumers' price response behavior identification}
\begin{figure}
\centering
{\includegraphics[width=0.45\textwidth]{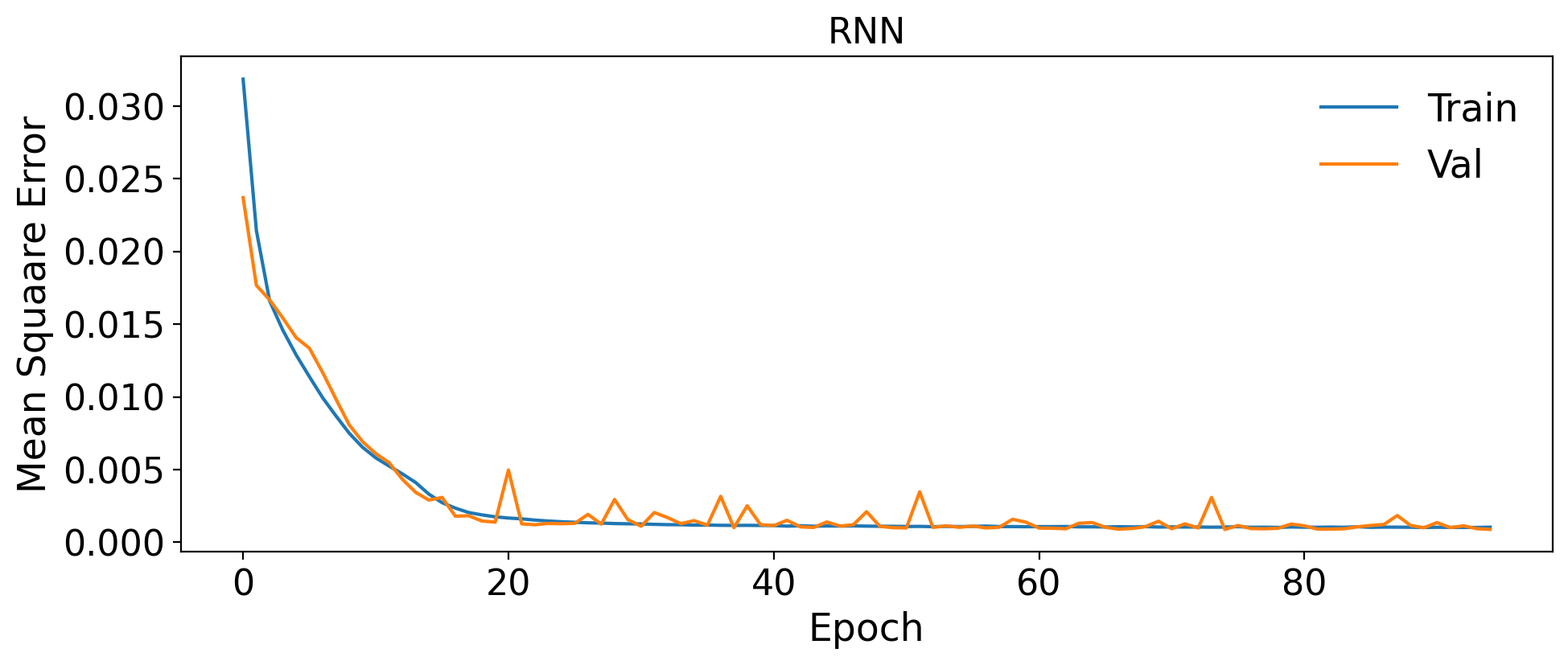}}
\caption{RNN training process.}
\label{train}
\vspace{-.4cm}
\end{figure}

In the RNN's training process, we set the learning rate to 0.0001, 2 hidden layers with 10 neurons, and chose Adam as the gradient-based optimizer. The first hidden layer uses ReLU as the activation function, and the second uses SeLU~\cite{selu}. The reason is that consumers' demand change is both positive and negative; ReLU will eliminate the negative part and kill the neurons when the negative appears. Thus, use SeLU to replace ReLU but keep the ReLU in the first layer to reduce the computation time. The training process is shown in Fig. \ref{train}, which presents good convergence of RNN. Although some minor fluctuations exist, there is nothing overfitting phenomenon, and MSE keeps around 0.0005. Besides, the prediction error almost concentrates on 0, indicating high prediction accuracy. Here we need to train 10 RNNs to get the price response behaviors of each group. The overall training time is 200s, and the RMSE and MAPE for each group remain at a low level, e.g., 0.065 and 0.78 in group 7. 

\begin{figure}[htbp]
    \centering{
    \includegraphics[width=0.45\textwidth]{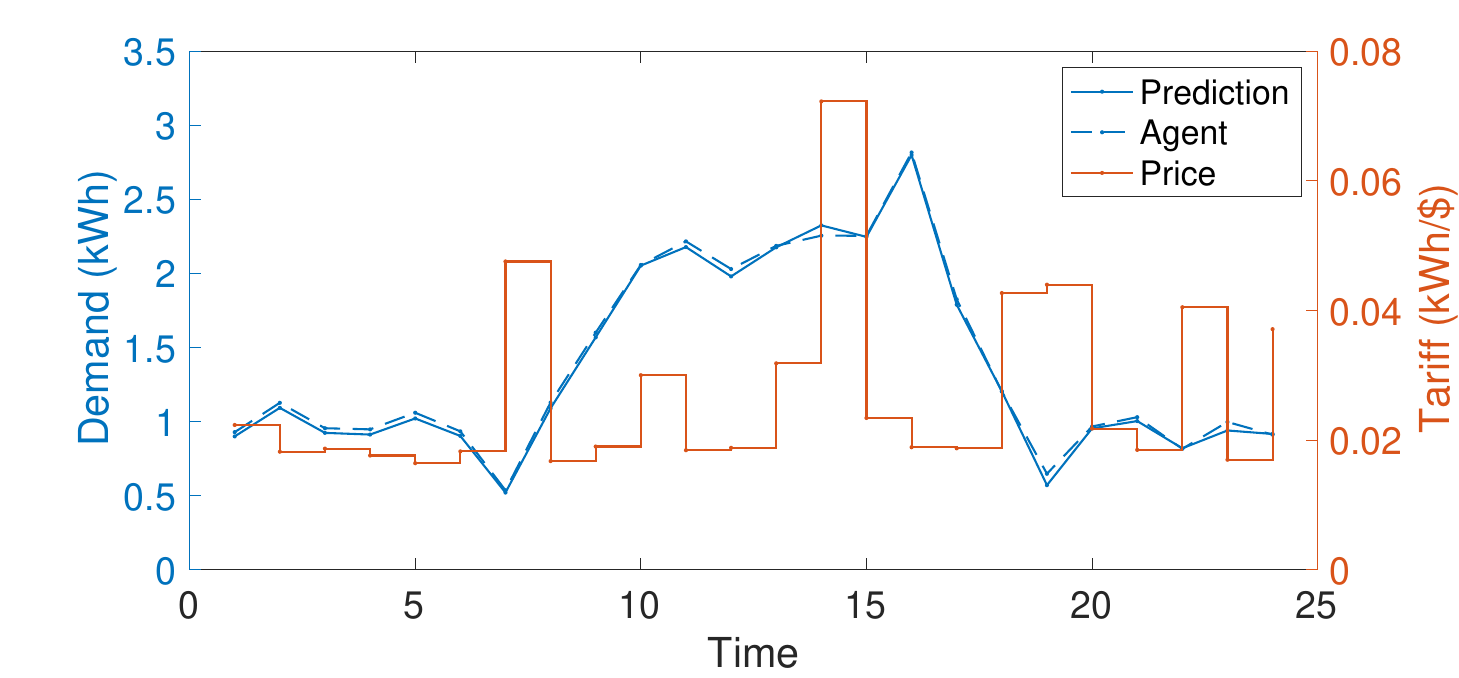}}
    \caption{Demand identification results of group 5 with tariffs.}
    \label{predict}
    \vspace{-.2cm}
\end{figure}

By training the RNN architecture using the provided dataset, we are able to derive network parameters that effectively capture consumers' price response behaviors, as represented by the agent model. This is clearly demonstrated in Fig. \ref{predict}, where the prediction results closely match the behavior depicted by the agent model.

\subsection{Time varying tariff results}
\subsubsection{Case (1) tariff design}
\begin{figure}
\centering
{\includegraphics[width=0.45\textwidth]{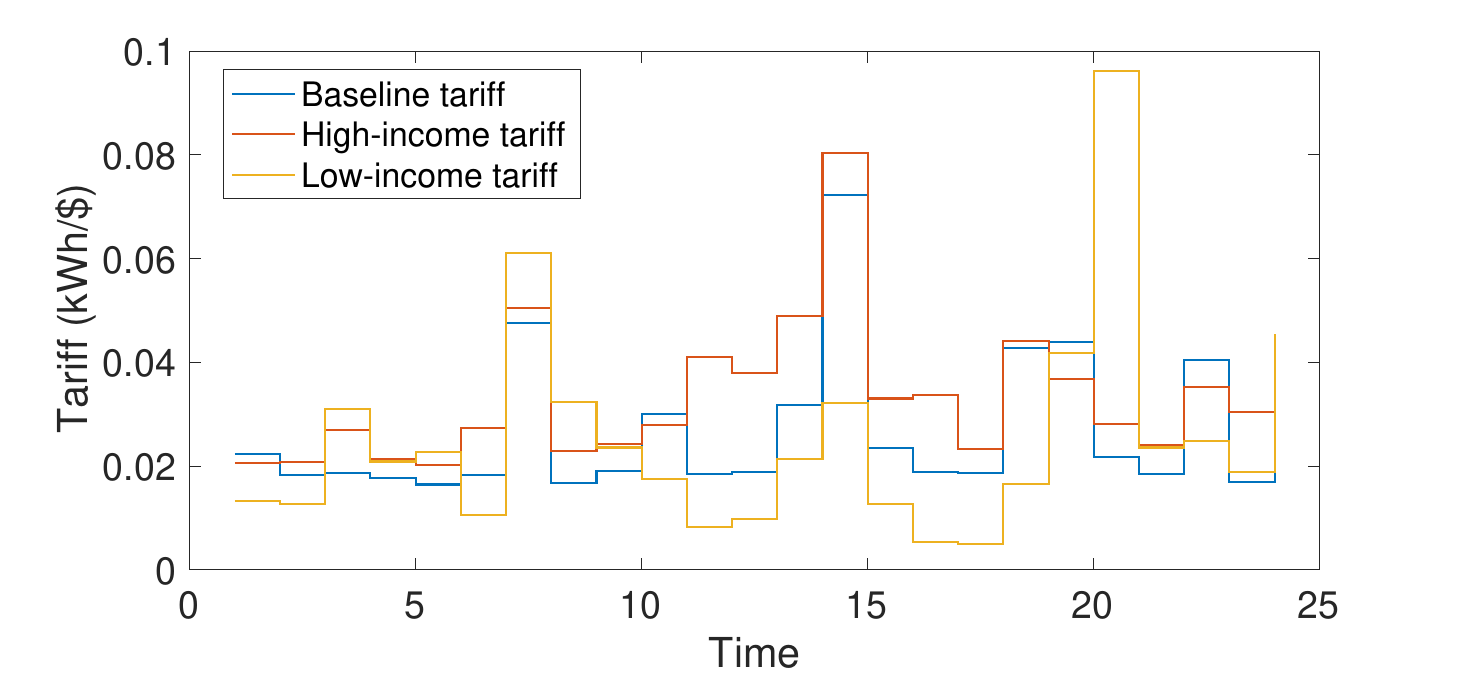}}
\caption{Modified tariffs of low-income group 1 and high-income group 9.}
\label{1price}
\vspace{-.4cm}
\end{figure}

\begin{figure*}
\centering
\subfigure[Baseline tariff]{\includegraphics[width=0.3\textwidth]{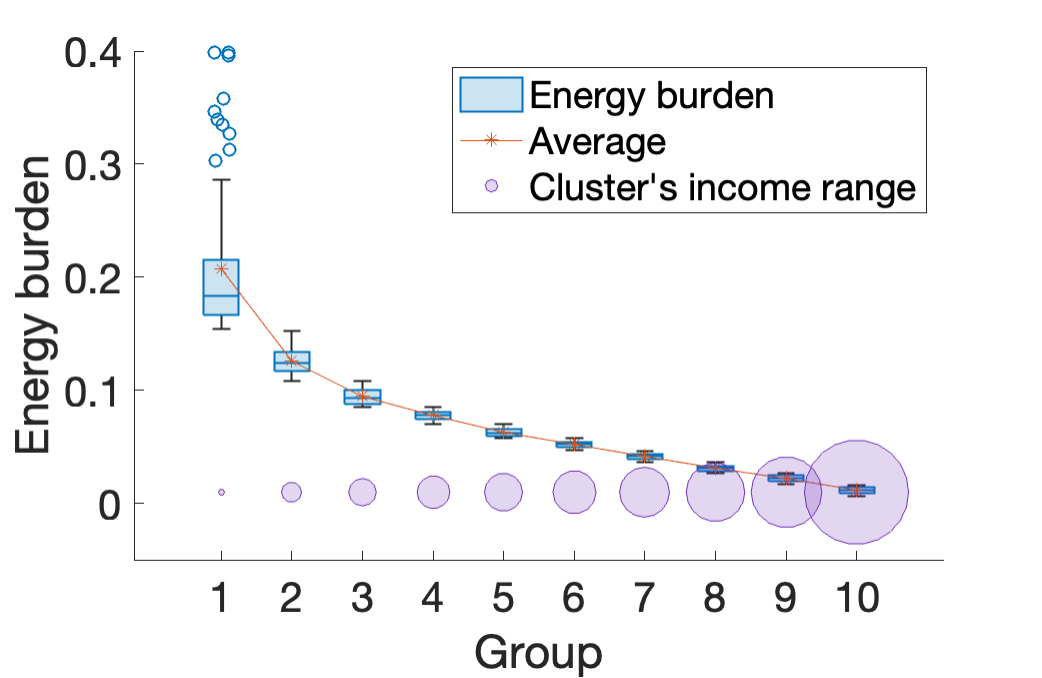}}
\subfigure[Proposed tariff]{\includegraphics[width=0.3\textwidth]{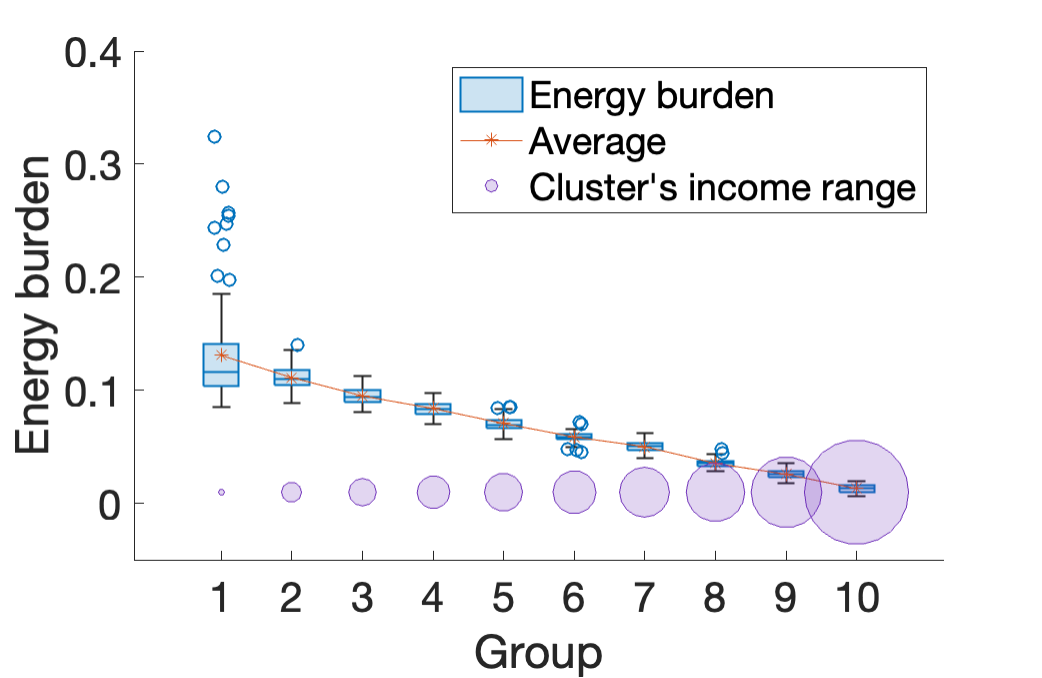}}
\subfigure[Proposed tariff with DR events]{\includegraphics[width=0.3\textwidth]{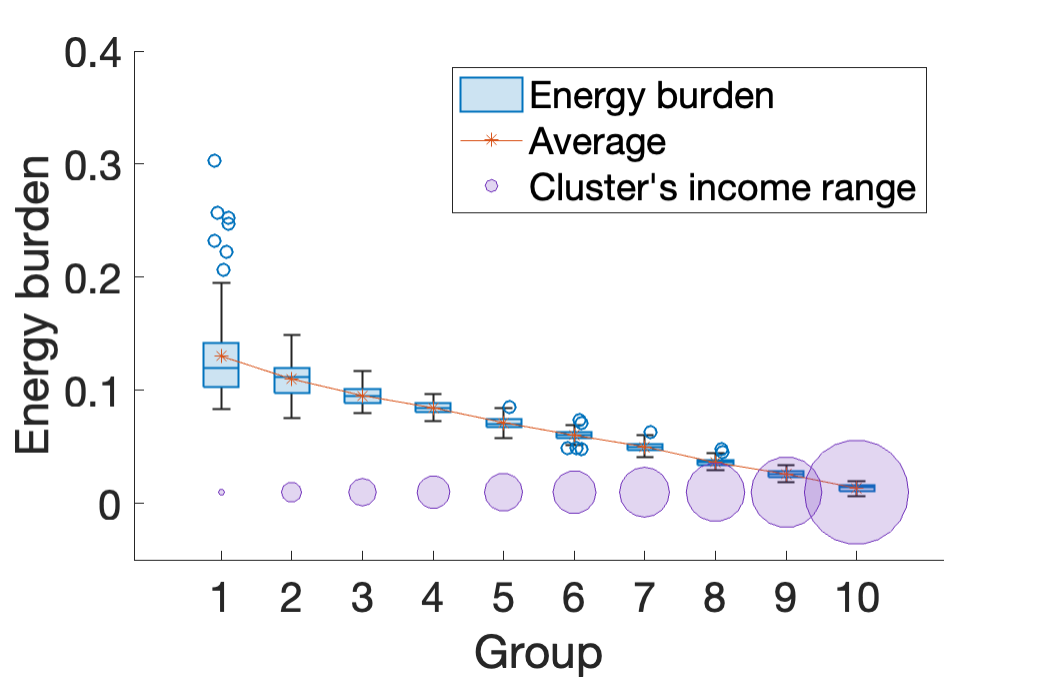}}
\caption{Consumers' energy burden under different tariffs.}
\label{1burden}
\vspace{-.4cm}
\end{figure*}

In the tariffs design case, the primary objective is to optimize consumers' load arrangement by incentivizing the reduction and shifting of load, while also redistributing the electricity costs among consumers of varying income levels. Fig. \ref{1price} shows the modified tariffs compared with the baseline scenario. The overall tariffs for low-income groups are reduced while those for high-income groups are increased, particularly during peak demand time slots 11-15. Besides, the high tariffs are shifted to demand valley time slots. As a result of these modified tariffs, consumers adjust parts of reducible demand motivated by the tariffs to reduce the electricity bills and energy burden. This is further illustrated in Fig. \ref{1burden}, which shows energy burden in both the baseline and modified scenario.  

From Fig. \ref{1burden}, it is evident that the energy burden experienced by low-income groups (1-4) shows a reduction in terms of variation, range, and expected value. Conversely, high-income groups (5-10) witness an increase in their energy burden due to cost sharing. Moreover, the aggregated demand under the modified tariff is 32,725 kWh, which is similar to the baseline tariff's 35,255 kWh. However, the modified tariff achieves a more favorable energy burden distribution, particularly benefiting low-income consumers by reallocating electricity costs based on income level and demand conditions. The analysis also shows that consumers with higher energy burdens undergo greater modifications and receive increased benefits. In addition to value, the range of each group also expands, suggesting that wealthier consumers contribute a larger share of the costs to promote energy equity. Additionally, the modified average line exhibits a more linear trend, reflecting a fair approach to redistributing electricity bills.

The computation time for this case is about 81s, including 41s for tariffs optimization and 40s for demand optimization using the agent model after receiving the modified tariffs. Actually, the computation time does not increase with the number of consumers, because tariff optimization is performed at the group level and demand optimization is a separate process for each consumer, which can be efficiently achieved through multi-core parallel computing. Besides, the income of the utility increased by nearly 45\$, accounting for 4.5\% when compared to the baseline profit. This increment falls within an acceptable range, demonstrating the viability and effectiveness of the proposed approach.

\subsubsection{Case (2) DR event}
\begin{figure}
\centering
{\includegraphics[width=0.45\textwidth]{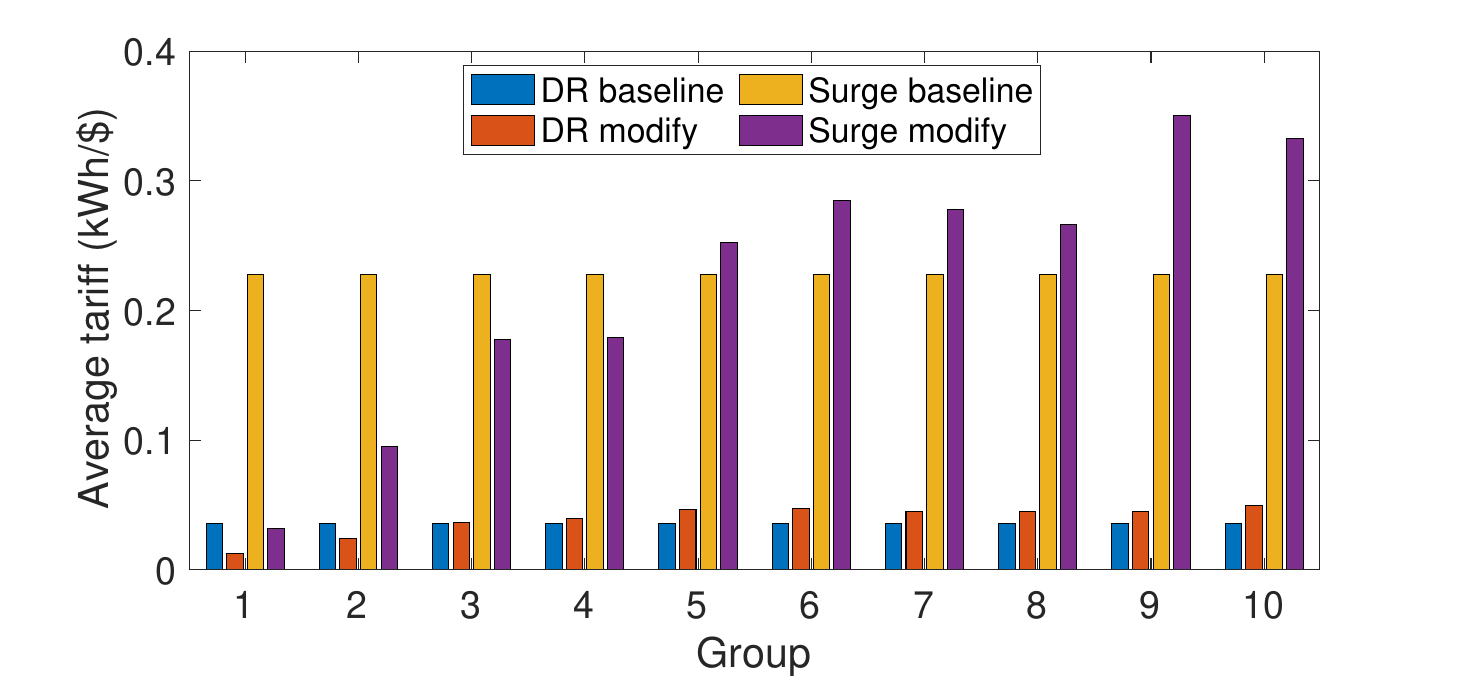}}
\caption{Modified tariff in DR and price surge case.}
\label{2price}
\vspace{-.4cm}
\end{figure}

In this case, we apply our method with a 2\% demand reduction target at the highest 4 time slots. Fig. \ref{2price} shows the average tariff in peak time slots.
The modified tariffs for high-income groups increase to motivate high-income consumers to reduce demand, while tariffs for low-income groups reduce or keep at a low level even during these peak time slots. 
This shows that high-income consumers take the risk during demand reduction events, protect low-income consumers, and share part of the cost of low-income consumers to facilitate energy equity. 


The modified energy burden, as depicted in Fig. \ref{1burden}, demonstrates a significant decrease in energy burden for low-income groups, while high-income groups experience an increase, indicating that they take risks. These findings highlight the effectiveness of our approach in achieving a fair allocation of electricity costs, without compromising the effectiveness of DR measures and protecting low-income consumers from excessive demand reductions. Besides, high-income consumers' energy burden is slightly higher than in case (1), showing that high-income consumers respond more to the DR.


The computation time for this case is 140s, comprising 100s for price optimization and 40s for demand optimization using the individual consumers' agent model. This duration is longer than the tariff design case because the DR constraints prolong the tariff optimization process. The utility's income increased by 55\$ in this case, accounting for 5.5\% compared to the baseline profit. The reason is that the DR event requires consumers to reduce more demand, prompting the utility to set higher prices and get additional revenue.

\subsubsection{Case (3) price surge}
In this case, we intentionally create significantly higher wholesale market prices by multiplying the price by 5 for the 14th and 16th time slots, as depicted in the Surge base and Surge scenarios in Fig. \ref{2price}. The average modified tariffs for these time slots clearly demonstrate that the burden of tariff surges falls on high-income consumers as their tariffs increase, while low-income consumers continue to receive lower tariffs and maintain similar load consumption patterns. It is worth noting that the modified tariffs are also influenced by demand, particularly for high-income groups. For instance, group 9 exhibits a slightly higher tariff than Group 10 due to its higher demand (2.8kWh) compared to Group 10's demand (2.6kWh). This tariff adjustment aligns with the utility company's revenue adequacy constraint, where charging higher tariffs to high-demand groups ensures greater payments from consumers. 
The energy burden results show a slight increase for high-income groups compared to other scenarios, as they bear the impact of the price surges to protect low-income consumers. 

The computation time for this case is 86s, with 46s for price optimization and 40s for demand optimization. The utility's revenue increased by 75\$ (5.4\%), similar to other cases. The slightly higher revenue can be attributed to the changes in wholesale market prices and baseline demand.

\subsection{Model validation and reliability analysis}
\subsubsection{Model validation}
\begin{figure}
\centering
\subfigure[Utility's revenue comparison]{\includegraphics[width=0.24\textwidth]{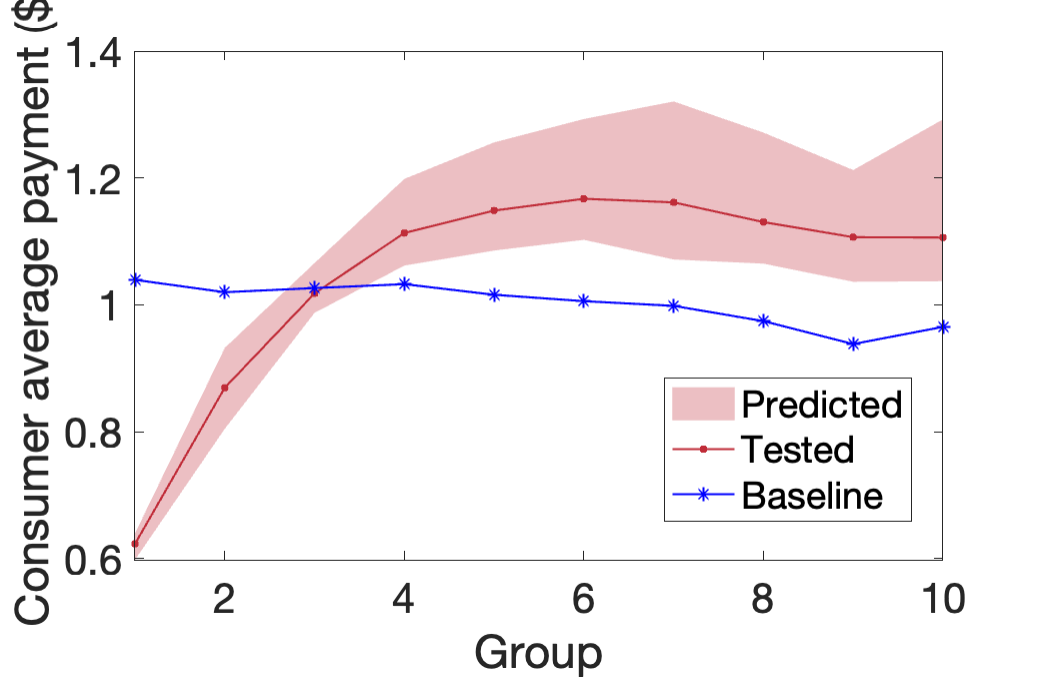}}
\subfigure[Energy burden comparison]{\includegraphics[width=0.24\textwidth]{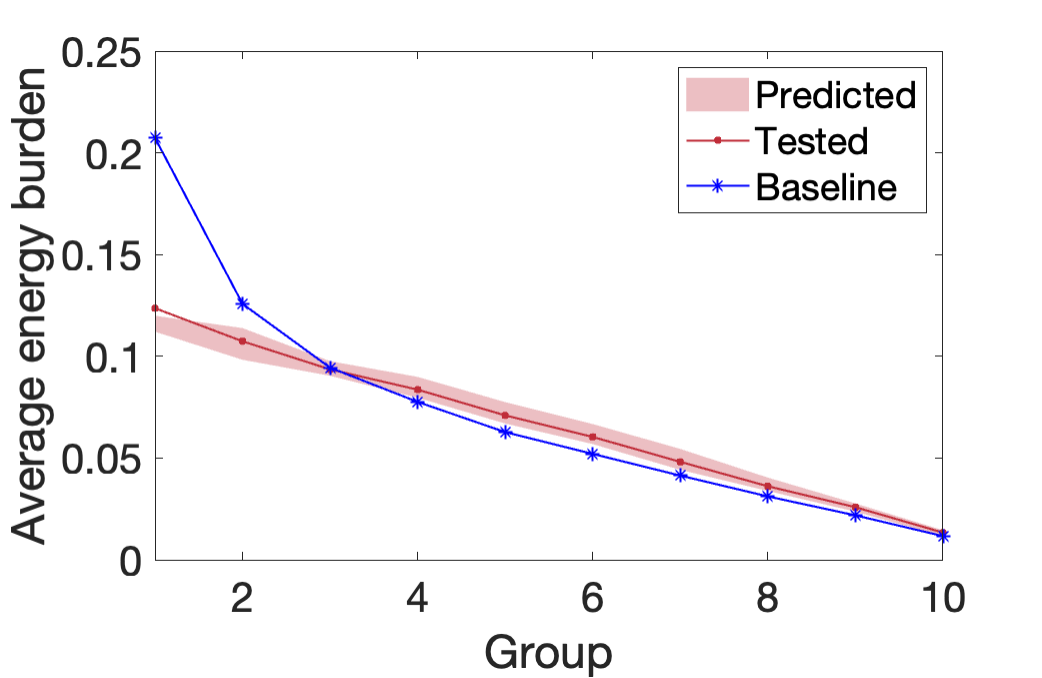}}
\caption{Model validation with consumers' individual agent model.}
\label{benchmark}
\vspace{-.4cm}
\end{figure}

To validate the proposed approach, we use the agent model to simulate real consumers and generate the tested scenario. The results, including the energy burden, utility revenue, and peak shaving, are depicted in Fig. \ref{benchmark} for the baseline, tested, and predicted scenarios with the modified tariff. The profit and the energy burden in each group consistently outperform those of the baseline scenario, demonstrating the effectiveness of our methods. The energy burden of the middle group shows a slight increase, which can be attributed to the consideration of both income and demand information, with the middle group exhibiting higher demand. Furthermore, small changes in the energy burden of higher groups actually corresponded to larger shifts in payments.

There are minor differences observed between tested and predicted scenarios, which mainly come from two parts, the prediction error of the RNN and the grouping approach. In the identification process, the RNN learns the price response behavior of each group based on the average demand of the entire group, instead of individual consumers' demand. However, the tested scenario is calculated by each consumer's agent model. Although the groups can approximate consumers' behavior to some extent, differences still exist when compared to the behavior of individual tested consumers. Specifically, we initialize the RNN training 23 times, the comparison results show the energy burden difference is less than 0.015 (within 11\% when compared to the tested scenario), and the profit difference is within 0.088\$ (8\%). In terms of peak shaving, the proposed method realizes a 2\% demand reduction in peak time slots. The demand reduction in the tested scenario is 18.2\%, 2.4\%, 7.4\%, 2.3\%, similar to that of the predicted scenarios with 15.8\%, 3.9\%, 8.9\%, 4.6\%.

\subsubsection{Demand reduction reliability}
\begin{figure}[htbp]
    \centering{
    \includegraphics[width=0.45\textwidth]{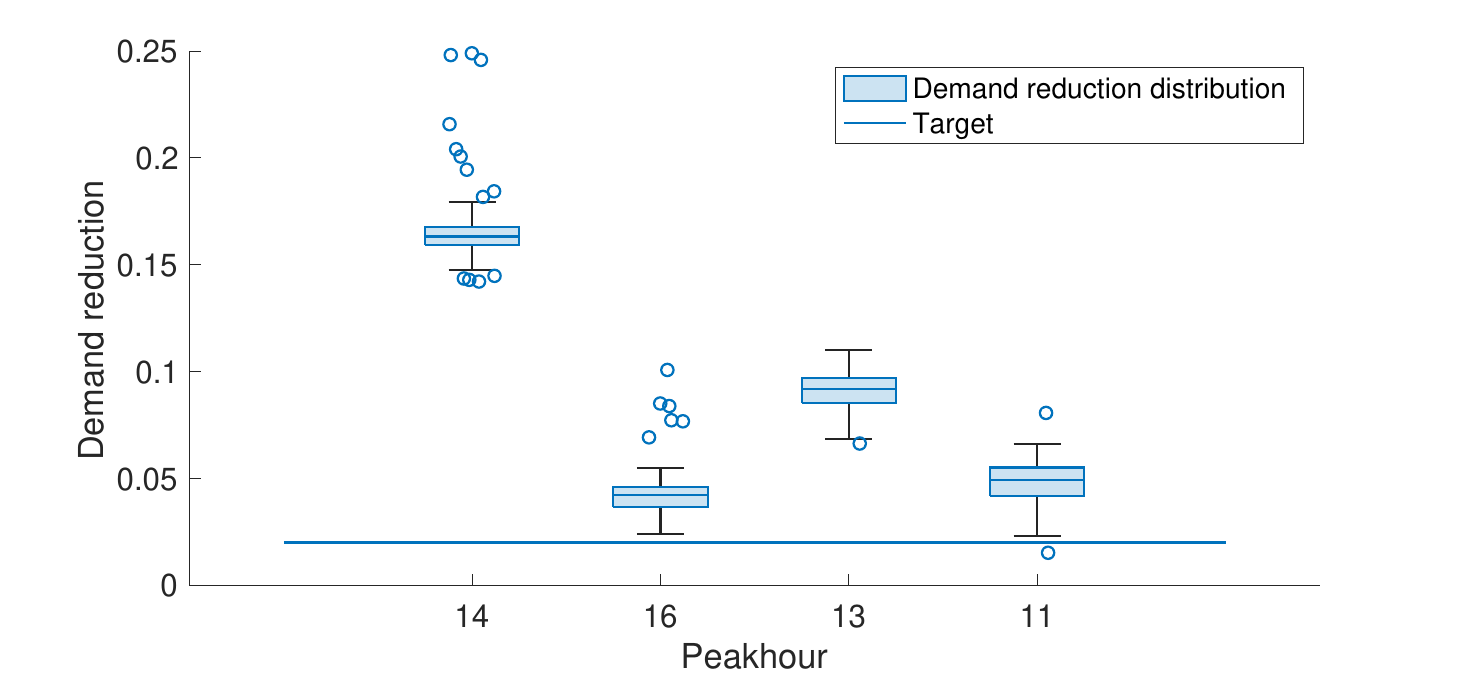}}
    \caption{Demand reduction distribution considering prediction errors.}
    \label{reliable}
    \vspace{-.2cm}
\end{figure}

We analyze the influence of RNN's prediction error (Fig. \ref{train}) on the demand reduction goal. Through the validation process of RNN, we get the prediction error distribution, as shown in Fig. \ref{reliable}. It is obvious that although the demand reduction ratio is differently affected by the prediction error, the peak demand reduction targets can be achieved with success rates of 100\%, 100\%, 100\%, and 99.6\%, indicating the strong reliability of our method. 

\section{Conclusion}
This paper proposes an energy equity pricing tariff design method for responsive consumers, using RNN to capture consumers' time-coupled price response behaviors and consider social demographics in the formulations. By leveraging a dataset constructed through the consumers' agent model, we obtain consumers' personal price response behaviors and use it to design energy equity pricing schemes. We apply our method to tariff design, DR event, and price surge cases with demand reduction constraints and high-price time slots, respectively. The results show our method effectively reduces consumers' energy burden difference and protects low-income consumers from the adverse effects of price surges. The demand reduction targets also show great robustness when considering the effect of prediction error. By testing our method on  agent-based consumer models, the identification and optimization framework results are consistent with the benchmark, demonstrating that our method accurately captures consumers' price response behavior and works well in real conditions.
Future directions will move toward distributed energy resource conditions and enhancing the identification process by including more detailed methods, e.g., building uncertainty blocks in structured neural networks.
\vspace{-.2cm}
\appendix
\subsection{Agent model}
It is noted that many agent models for consumers' price response process can be used in work. Here we use the following agent model to simulate consumers' price response process:
\begin{align}
    \min \limits_{\bm{D}_{\mathrm{r}},\bm{D}_{\mathrm{s}}} \ & \bm{p}^{T}\bm{D} + c_{1} \bm{D}_{\mathrm{r}}^2 + c_{2} \bm{D}_{\mathrm{s}}^2 \\
    &\bm{D} = \bm{D}_{0}+ \bm{D}_{\mathrm{r}} + \bm{D}_{\mathrm{s}} \\
    \mathrm{s.t.} \ &\sum_{t \in T} D_{\mathrm{s},t} = 0  \\
    &\underline{D_{\mathrm{s}}} \leq \bm{D}_{\mathrm{s}} \leq \overline{D_{\mathrm{s}}} \\
    &\underline{D_{\mathrm{r}}} \leq \bm{D}_{\mathrm{r}} \leq \overline{D_{\mathrm{r}}}
\end{align}
where $c_{1}, c_{2}$ are the parameters of reduced demand and shiftable demand, which is determined by each consumer's price response behaviors; $\underline{D_{\mathrm{s}}}, \overline{D_{\mathrm{s}}}$ are the lower and upper bond of shiftable demand, respectively; $\underline{D_{\mathrm{r}}}, \overline{D_{\mathrm{r}}}$ are the lower and upper bond of reduced demand, respectively.

\subsection{Convergence proof of proposed method}
We first show the derivative of (\ref{newobj}) with regard to $p$ is similar to the derivative with regard to $\bm{W}$.
\begin{align}
     &\frac{\partial D_{t}}{\partial p_{t}}
     = \frac{\partial D_{t}}{\partial \bm{h}_{t}^{(L)}}  
    \prod_{l=2}^{L} \left( \frac{\partial \bm{h}_{t}^{(l)}}{\partial \bm{h}_{t}^{(l-1)}} 
     + \frac{\partial \bm{h}_{t}^{(l)}}{\partial \bm{h}_{t-1}^{(l)}} 
    \right)
    \left( \frac{\partial \bm{h}_{t}^{(1)}}{\partial p_{t}} + \frac{\partial \bm{h}_{t}^{(1)}}{\partial \bm{h}_{t-1}^{(1)}} \right)  \label{proof1}
\end{align}
\vspace{-0.3cm}
\begin{align}
\begin{split}
    \frac{\partial D_{t}}{\partial \bm{W}_{\mathrm{p}}^{l}} = \frac{\partial D_{t}}{\partial \bm{h}_{t}^{(L)}} 
     \prod_{j=l+1}^{L} &\left( \frac{\partial \bm{h}_{t}^{(j)}}{\partial \bm{h}_{t}^{(j-1)}} 
     + \frac{\partial \bm{h}_{t}^{(j)}}{\partial \bm{h}_{t-1}^{(j)}} 
    \right)  \\
    * &\left( \frac{\partial \bm{h}_{t}^{(l)}}{\partial \bm{W}_{\mathrm{p}}^{(l)}} + \frac{\partial \bm{h}_{t}^{(l)}}{\partial \bm{h}_{t-1}^{(l)}} \right)  \label{proof2} 
\end{split}
\end{align}

When $l=1$, (\ref{proof1}) and (\ref{proof2}) have the same structure, and when $l>1$, they also have similar structure while (\ref{proof2}) has short derivative chain. The difference is that (\ref{proof1}) obtain the input data $p$, while (\ref{proof2}) get the neural network parameters $\bm{W}$. Then according to ~\cite{globalmini}, because the objective (\ref{obj}) is similar to the loss function defined in the training of the neural network, i.e., both are the distance to constant vectors expressed by $l2$ norm, we formulate the following dynamics to express predictions:
\begin{align}
    E_{t}(k)&=\gamma f(\theta(k),p_{t})\\
    \frac{\partial \bm{E}(k)}{\partial k} &= \bm{G}(\overline{E}-\bm{E}(k))    \label{dynamic}
\end{align}
where $\gamma$ are parameters to transfer $D$ to $E$; $\theta$ is neural network parameters; $k$ is the index of gradient descent iteration; $\bm{G}$ is the Gram matrix, which is defined as follows:
\begin{align}
\begin{split}
    \bm{G}_{i,j}(k)&=\left \langle \frac{\partial E_{i}(k)}{\partial \bm{p}(k)} , \frac{\partial E_{j}(k)}{\partial \bm{p}(k)} \right \rangle \\
    &=\left \langle \frac{\partial E_{i}}{\partial \bm{D}} \frac{\partial \bm{D}}{\partial \bm{p}} , \frac{\partial E_{j}}{\partial \bm{D}} \frac{\partial \bm{D}}{\partial \bm{p}} \right \rangle
    + \left \langle \frac{\partial E_{i}}{\partial \bm{D}} \frac{\partial \bm{D}}{\partial \bm{p}} , \frac{\partial E_{j}}{\partial \bm{p}}  \right \rangle \\
    &+ \left \langle \frac{\partial E_{j}}{\partial \bm{D}} \frac{\partial \bm{D}}{\partial \bm{p}} , \frac{\partial E_{i}}{\partial \bm{p}}  \right \rangle
     + \left \langle \frac{\partial E_{i}}{\partial \bm{p}}  , \frac{\partial E_{j}}{\partial \bm{p}}  \right \rangle
\end{split}
\end{align}

The other part is the linear distance between $D$ and $D_{0}$, which also satisfies similar dynamics as shown in (\ref{dynamic}). 

According to~\cite{globalmini}, if $G(k)$ is strictly positive definite, then the loss $(\bm{E} - \overline{E})^T (\bm{E} -\overline{E})$ decreases at the $k^{\mathrm{th}}$ iteration according to the analysis of power method if the learning rate $\eta$ is properly chosen. Each element of $\bm{G}(k)$ is positively defined as it is formulated by the inner product, and the sum of the positively defined matrix is still positively defined. Thus, $\bm{G}(k)$ is positive definite, and based on the dynamics, the objective function will decrease as the iterations go on, showing that the gradient-based solution method can converge after finite iterations as long as the learning rate is sufficiently low.



%




\bibliographystyle{IEEEtran}
\bibliography{Ref}

\begin{thebibliography}{10}
\providecommand{\url}[1]{#1}
\csname url@samestyle\endcsname
\providecommand{\newblock}{\relax}
\providecommand{\bibinfo}[2]{#2}
\providecommand{\BIBentrySTDinterwordspacing}{\spaceskip=0pt\relax}
\providecommand{\BIBentryALTinterwordstretchfactor}{4}
\providecommand{\BIBentryALTinterwordspacing}{\spaceskip=\fontdimen2\font plus
\BIBentryALTinterwordstretchfactor\fontdimen3\font minus
  \fontdimen4\font\relax}
\providecommand{\BIBforeignlanguage}[2]{{%
\expandafter\ifx\csname l@#1\endcsname\relax
\typeout{** WARNING: IEEEtran.bst: No hyphenation pattern has been}%
\typeout{** loaded for the language `#1'. Using the pattern for}%
\typeout{** the default language instead.}%
\else
\language=\csname l@#1\endcsname
\fi
#2}}
\providecommand{\BIBdecl}{\relax}
\BIBdecl

\bibitem{energy}
IEA, ``World energy outlook. 2022,''
  \url{https://www.iea.org/reports/world-energy-outlook-2022}.

\bibitem{qdr2006benefits}
Q.~Qdr, ``Benefits of demand response in electricity markets and
  recommendations for achieving them,'' \emph{US Dept. Energy, Washington, DC,
  USA, Tech. Rep}, vol. 2006, 2006.

\bibitem{el2020contracted}
R.~El~Geneidy and B.~Howard, ``Contracted energy flexibility characteristics of
  communities: Analysis of a control strategy for demand response,''
  \emph{Applied Energy}, vol. 263, p. 114600, 2020.

\bibitem{review}
J.~S. Vardakas, N.~Zorba, and C.~V. Verikoukis, ``A survey on demand response
  programs in smart grids: Pricing methods and optimization algorithms,''
  \emph{IEEE Communications Surveys \& Tutorials}, vol.~17, no.~1, pp.
  152--178, 2015.

\bibitem{tou3}
``Low carbon london project,'' \url{https://innovation.ukpowernetworks.co.uk
  /projects/low-carbon-london/}.

\bibitem{tou9}
S.~S. George, M.~Sullivan, J.~Potter, and A.~Savage, ``{Time-of-Use Pricing
  Opt-In Pilot Plan},'' California public utilities commission, Tech. Rep., 12
  2015.

\bibitem{tou4}
``Con edison time-of-use rates,''
  \url{https://www.coned.com/en/accounts-billing/your-bill/time-of-use}.

\bibitem{tou5}
``Srp time-of-use price plan™ (tou),''
  \url{https://www.srpnet.com/price-plans/residential-electric/time-of-use}.

\bibitem{toupeak}
P.~Alexeenko and E.~Bitar, ``Achieving reliable coordination of residential
  plug-in electric vehicle charging: A pilot study,'' \emph{Transportation
  Research Part D: Transport and Environment}, vol. 118, p. 103658, 2023.

\bibitem{tou8}
A.~Madduri, M.~Foudeh, P.~Phillips, A.~Gupta, J.~Hsu, A.~Jain, P.~Voris,
  J.~Lamming, and A.~Magie, ``{Advanced Demand Flexibility Management},''
  California public utilities commission, Tech. Rep., 01 2022.

\bibitem{rtp5}
``Power smart pricing plan,'' \url{https://www.ameren.com/illinois/account/
  customer-service/bill/power-smart-pricing/}.

\bibitem{rtp6}
``Ercot day-ahead and real-time market prices,''
  \url{https://www.ercot.com/mktinfo/prices}.

\bibitem{nyt}
\BIBentryALTinterwordspacing
M.~N. d.~R. Giulia, B.-B. Nicholas, and P.~Ivan, ``His lights stayed on during
  texas’ storm. now he owes \$16,752.'' [Online]. Available:
  \url{https://www.nytimes.com/2021/02/20/us/texas-storm-electric-bills.html\#commentsContainer}
\BIBentrySTDinterwordspacing

\bibitem{tou1}
T.~Schittekatte, D.~S. Mallapragada, P.~L. Joskow, and R.~Schmalensee,
  ``Electricity retail rate design in a decarbonized economy: An analysis of
  time-of-use and critical peak pricing,'' National Bureau of Economic
  Research, Tech. Rep., 2022.

\bibitem{tou6}
R.~Li, Z.~Wang, C.~Gu, F.~Li, and H.~Wu, ``A novel time-of-use tariff design
  based on gaussian mixture model,'' \emph{Applied Energy}, vol. 162, pp.
  1530--1536, 2016.

\bibitem{tou7}
V.~Venizelou, N.~Philippou, M.~Hadjipanayi, G.~Makrides, V.~Efthymiou, and
  G.~E. Georghiou, ``Development of a novel time-of-use tariff algorithm for
  residential prosumer price-based demand side management,'' \emph{Energy},
  vol. 142, pp. 633--646, 2018.

\bibitem{rtp2}
K.~B. Herter and G.~Situ, ``{Analysis of Potential Amendments to the Load
  Management Standards},'' California public utilities commission, Tech. Rep.,
  12 2021.

\bibitem{rtp3}
C.~P.~U. Commission, ``{Distributed Energy Resources Action Plan 2.0},''
  \url{https://www.cpuc.ca.gov/about-cpuc/divisions/energy-division/der-action-plan},
  California public utilities commission, Tech. Rep., 4 2022.

\bibitem{rtp4}
M.~Hogan, ``{Employing Price-Responsive Demand to Reduce the Investment
  Challenge},''
  \url{https://www.esig.energy/price-responsive-demand-to-reduce-investment-challenge/},
  Energy systems Integration Group, Tech. Rep., 1 2023.

\bibitem{rtp7}
L.~He, Y.~Liu, and J.~Zhang, ``An occupancy-informed customized price design
  for consumers: A stackelberg game approach,'' \emph{IEEE Transactions on
  Smart Grid}, vol.~13, no.~3, pp. 1988--1999, 2022.

\bibitem{rtp8}
N.~Liu, X.~Yu, C.~Wang, C.~Li, L.~Ma, and J.~Lei, ``Energy-sharing model with
  price-based demand response for microgrids of peer-to-peer prosumers,''
  \emph{IEEE Transactions on Power Systems}, vol.~32, no.~5, pp. 3569--3583,
  2017.

\bibitem{rtp9}
K.~Khezeli and E.~Bitar, ``Risk-sensitive learning and pricing for demand
  response,'' \emph{IEEE Transactions on Smart Grid}, vol.~9, no.~6, pp.
  6000--6007, 2018.

\bibitem{rtp10}
T.~M. Aljohani, A.~F. Ebrahim, and O.~A. Mohammed, ``Dynamic real-time pricing
  mechanism for electric vehicles charging considering optimal microgrids
  energy management system,'' \emph{IEEE Transactions on Industry
  Applications}, vol.~57, no.~5, pp. 5372--5381, 2021.

\bibitem{rtp11}
A.-H. Mohsenian-Rad and A.~Leon-Garcia, ``Optimal residential load control with
  price prediction in real-time electricity pricing environments,'' \emph{IEEE
  Transactions on Smart Grid}, vol.~1, no.~2, pp. 120--133, 2010.

\bibitem{ATUS}
\BIBentryALTinterwordspacing
``The american time use survey (atus).'' [Online]. Available:
  \url{https://www.bls.gov/tus/}
\BIBentrySTDinterwordspacing

\bibitem{add1}
L.~B.~T. Jill S.~Tietjen, Marija D.~Ilic and N.~N. Schulz, \emph{Women in
  Power: Research and Development Advances in Electric Power Systems}.\hskip
  1em plus 0.5em minus 0.4em\relax Springer Nature Switzerland AG, Cham,
  Switzerland., 2023.

\bibitem{add4}
X.~Jin, K.~Baker, S.~Isley, and D.~Christensen, ``User-preference-driven model
  predictive control of residential building loads and battery storage for
  demand response,'' in \emph{2017 American Control Conference (ACC)}.\hskip
  1em plus 0.5em minus 0.4em\relax IEEE, 2017, pp. 4147--4152.

\bibitem{Nat1}
R.~K. Jain, J.~Qin, and R.~Rajagopal, ``Data-driven planning of distributed
  energy resources amidst socio-technical complexities,'' \emph{Nature Energy},
  vol.~2, no.~8, pp. 1--11, 2017.

\bibitem{Chen}
L.~Chen, N.~Liu, C.~Li, and J.~Wang, ``Peer-to-peer energy sharing with social
  attributes: A stochastic leader--follower game approach,'' \emph{IEEE
  Transactions on Industrial Informatics}, vol.~17, no.~4, pp. 2545--2556,
  2020.

\bibitem{ml1}
M.~Dolányi, K.~Bruninx, J.-F. Toubeau, and E.~Delarue, ``Capturing electricity
  market dynamics in strategic market participation using neural network
  constrained optimization,'' \emph{IEEE Transactions on Power Systems}, pp.
  1--13, 2023.

\bibitem{ml4}
R.~Nellikkath and S.~Chatzivasileiadis, ``Physics-informed neural networks for
  ac optimal power flow,'' \emph{Electric Power Systems Research}, vol. 212, p.
  108412, 2022.

\bibitem{ml2}
Y.~Chen, Y.~Shi, and B.~Zhang, ``Modeling and optimization of complex building
  energy systems with deep neural networks,'' in \emph{2017 51st Asilomar
  Conference on Signals, Systems, and Computers}, 2017, pp. 1368--1373.

\bibitem{ml3}
J.~Drgo{\v{n}}a, A.~R. Tuor, V.~Chandan, and D.~L. Vrabie,
  ``Physics-constrained deep learning of multi-zone building thermal
  dynamics,'' \emph{Energy and Buildings}, vol. 243, p. 110992, 2021.

\bibitem{report_energyburden}
A.~Drehobl, R.~L., and A.~R., ``How high are household energy burdens,''
  Washington, DC: American Council for an Energy-Efficient Economy, Tech. Rep.,
  2020.

\bibitem{mitbook}
I.~Goodfellow, Y.~Bengio, and A.~Courville, \emph{Deep learning}.\hskip 1em
  plus 0.5em minus 0.4em\relax MIT press, 2016.

\bibitem{globalmini}
S.~Du, J.~Lee, H.~Li, L.~Wang, and X.~Zhai, ``Gradient descent finds global
  minima of deep neural networks,'' in \emph{International conference on
  machine learning}.\hskip 1em plus 0.5em minus 0.4em\relax PMLR, 2019, pp.
  1675--1685.

\bibitem{proof2}
Z.~Allen-Zhu, Y.~Li, and Z.~Song, ``On the convergence rate of training
  recurrent neural networks,'' \emph{Advances in neural information processing
  systems}, vol.~32, 2019.

\bibitem{proof4}
J.~Martens, \emph{Second-order optimization for neural networks}.\hskip 1em
  plus 0.5em minus 0.4em\relax University of Toronto (Canada), 2016.

\bibitem{selu}
G.~Klambauer, T.~Unterthiner, A.~Mayr, and S.~Hochreiter, ``Self-normalizing
  neural networks,'' \emph{Advances in neural information processing systems},
  vol.~30, 2017.

\end{thebibliography}

\ifCLASSOPTIONcaptionsoff
  \newpage
\fi

\end{document}